# Indonesian Social Media Sentiment Analysis with Sarcasm Detection

Edwin Lunando[#1], Ayu Purwarianti[#2]

*School of Electrical Engineering and Informatics, Institute Technology of Bandung,*
*10th Ganeca Street, Bandung, Indonesia.*
[1]edwinlunando@gmail.com
[2]ayu@stei.itb.ac.id

*Abstract*— **Sarcasm is considered one of the most difficult problem in sentiment analysis. In our ob-servation on Indonesian social media, for cer-tain topics, people tend to criticize something using sarcasm. Here, we proposed two additional features to detect sarcasm after a common sentiment analysis is conducted. The features are the negativity information and the number of interjection words. We also employed translated SentiWordNet in the sentiment classification. All the classifications were conducted with machine learning algorithms. The experimental results showed that the additional features are quite effective in the sarcasm detection.**

## I. Introduction

Recently, the topic of sentiment analysis, automatic classification of the opinion or sentiment conveyed by a text towards a subject, is highly researched. There are a lot of researches in this area which improves the technique in building a sentiment analysis system. Meanwhile, in the bussiness application industry, sentiment analysis application has been used comprehensively to give more information about the user insight about one topic to help them while making decision. Despite the high usage of the sentiment analysis application, there are still rooms for improvements. One of the problems that still become a challenge in sentiment analysis is sarcasm.

Sarcasm is using irony to mock or convey contempt. Sarcasm tranforms the polarity of the text into its opposite while, the text itself looked like the original sentiment. From our observation from 100 microblogging text with simple conversation topics such as food, life, and health, we found that sarcasm text is very rare. There were only 2 from 100 texts that contain sarcasm. But for sensitive topics such as government, brand, or politic, the quantity is greatly rose. There were 18 texts out of 100 texts containing sarcasm.

Sarcasm or irony is an old and well researched phenomenon in the field of linguistic and psychology [1]. Unfortunately, in the field of text mining or more specifically: sentiment analysis, detecting sarcasm automatically is still considered a difficult problem because lexical features do not give enough information to detect sarcasm. One approach to solve this problem was by employing lexical features such as unigram and pragmatic factors such as smileys or emoticons [5]. An interesting analysis from the field of lingustic tells us that lexical factors like presence of adjectives and adverbs, presence of interjections, and use of punctuations play a quite significant role in sarcasm[4]

In the basic sentiment analysis, many researches employed machine learning techniques such as Naïve Bayes, Maximum Entropy, and Support Vector machine because these algorithms are tend to outperform the other algorithm in the context of text classification [2][3][7]. Other than that, there is also a lexical resource, SentiWordNet [6], developed as a resource of word with sentiment score. These two techniques are also employed in our sentiment analysis system of Indonesian social media. This is different with other researches on sentiment analysis of Indonesian social media [2][3] which didn't address the sarcasm detection and didn't employ sentiment score in the SentiWordNet.

## II. Sentiment Analysis with Sarcasm Classification

An easy way to comply with the conference paper formatting requirements is to use this document as a template and simply type your text into it.

Similar with other sentiment analysis system, our sentiment classification system consists of preprocessing, fitur extraction and classification. The system architecture is as shown in Fig 1.

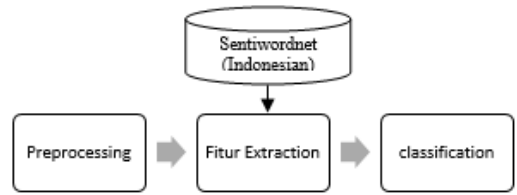

Fig 1. Architecture of the Sentiment Analysis System

### A. Preprocessing component

The preprocessing aims to minimize the vocabulary of terms used in the text message. In Indonesian text message of social media such as twitter or facebook, people tend to use slank words than the formal one such as using numeric to replace alphabet, repeating vocal characters, and using common informal words to replace the formal words [2]. To process such words, we employed the preprocessing such as follow:

1. Converse numeric character into alphabet, such as "ga2l" into "gagal" (fail)
2. Remove vocal repetition, such as "cemunguuuudh" into "cemungudh"
3. Translate informal words into formal words using dictionary, such as "cemungudh" into "semangat" (high spirit).

Here, even though some informal wordsare mispelled from formal words, but some other are completely different lexical than the formal words, therefore the strategy is to build a dictionary and use it to translate informal words into formal words.

### B. Feature extraction component

In the feature extraction, there are several features taken from the preprocessed text:

1. Unigram

We analyzed that unigram is more suitable for Indonesian social media text since the grammars used in Indonesian social media texts are various and informal.The unigram taken from the text is only the term that exists in our translated SentiWordNet. We translated English SentiWordNet into Indonesian using an available statistical machine translation (Google). In the translation, one Indonesian word with more than one English translation is given the average score of all the English translations.

For text "Pasangan Rieke dan Teten cocok untuk memajukan Jawa Barat!" (Rieke and Teten are suitable to bring forward West Java!), the unigram features are taken for "cocok" (suitable) and "memajukan" (bring forward) which gives SentiWordNet score of 0.5 and 0.375.

There are several phenomena in using unigram:

a. Negation

Negation words (such as "no" or "not") tend to change sentiment score of a certain word. For example, in the text "Televisi jaman sekarang tidak begitu bagus karena mahal" (nowadays televisions are notgood because they are expensive), the sentiment word "bagus" (good) is changed into negative because it is preceeded by a negation word ("tidak" or "not"). The negation words usually reside before the sentiment word and it can be located two or three words away from the sentiment word. We handled this by multiplying the score of sentiment word closest to the negation word by 1.

b. Word context

One word sentiment may change depends on its word context. For example, the word "mahasiswa" (student) is basically a neutral word but if this word is preceeded by "harga" (price) which makes it into "harga mahasiswa" (low price) then the word becomes a possitive word. To handle this problem, we used a special list of word such as used in [2].

c. Affix

Word with different affix may have different sentiment. For example "murah" (low) has a possitive sentiment, while "murahan" (twopenny) has a negative sentiment. Since not all suffix "an" have negative sentiment, we handled this problem by using a list of word such as used in [2].

2. Negativity

This feature represents the percentage of the negative sentiment in the topic of the text message. This feature is intended to catch global information. It gives information about the real sentiment of a certain topic. In order to get this feature, the topic of the text message should be extracted first. In this research, we did the topic extraction manually.

For example, in a topic about Indonesian singer, "Rhoma Irama wants to be Indonesia's President" has 80% negative sentiment. Then, the negativity value is 80% or 0.8.

3. Number of interjection words

This feature shows the number of interjection words from the text message. Example of interjection words are "aha", "bah", "nah", "wew", "wow", "yay", "uh", etc. We employ this feature based on our observation that among 100 sarcasm text, there are 20 text with interjection words. Below are the examples of sarcasm text with interjection.

| "Wow kk wow. hebat banget rhoma irama berani nyapres" |
|---|
| "jir, Polri Indonesia makin jago aja. Kapolda sendiri buron gitu." |
| "wah… cantik sekali cara telkomsel melayani pelanggannya, dikacangin coy." |

4. Question word

This feature is used to classify neutral text. By detecting the question word like "who", "what", "when", "how", "where", and "why" it will show that the text has no sentiment value. Almost all text with question words are classified into neutral text. This feature wotks like boolean value. If there are question word at a text, this feature will get "true" and vice versa.

### C. Classification component

In the classification component, there are two classification steps. The first classification is to classify each text into three sentiment classes: possitive, negative and neutral. The second classification is to classify the sarcasm of the possitive text. The flow is as shown in Fig 2.

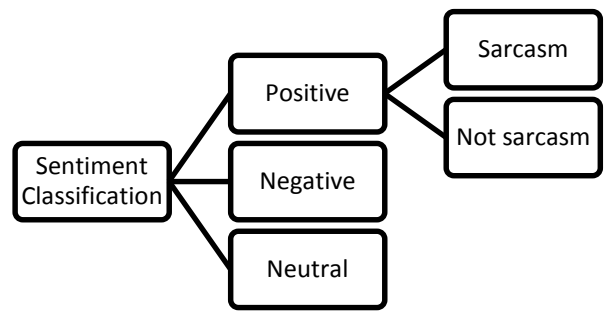

Fig 2. Classification Flow in the Sentiment Analysis System

All the classifications are conducted using several machine learning algorithms such as Naïve Bayes, Maximum Entropy, and Support Vector Machine. These algorithms were chosen because they have shown good accuracy in many text classification task [7].

In the first classification, the feature is the unigram, while in the second classification, the features are unigram, negativity and the number of interjection words.

Fig 3. Leveled Method in Sentiment Analysis

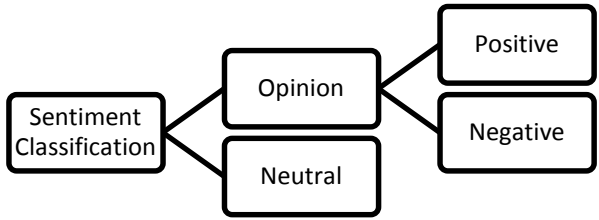

Fig 3 describe one of the classification method in sentiment analysis. Rather than directly classify a text into three classes, at first, this method classify a text into opinion and neutral text. After that, the opinion text will be classified into positive or negative class. [3] has tried to use this method, but they haven't compare those two methods.

III. EXPERIMENTS

*A. Experimental data*

The data used in the experiment was gathered manually from Twitter. The size of training data is 980 which consist of 502 neutral texts, 250 possitive texts and 228 negative texts. The size of testing data is 300 which consist of 200 neutral texts, 60 possitive texts and 40 negative texts. The training and testing data were collected from various topics such as politic, food, movie, and public figure.

In order to get the list of the negativity value, we generated it by using the search API from Twitter. We gathered 100 tweets from each topic and labeled the sentiment manually to get the number of positive and negative tweets.

*B. Experimental result and analysis*

There were three experiments conducted. The first experiment is on the first classification step, and the second experiment is on the second classification step(classification method), and the third classification is on the third classification step(sarcasm classification).

*1) Experiments on Sentiment Score*

As the first experiment, we evaluated the usage of the translated SentiWordNet in the first classification type which classify each text into 3 classes: possitive, negative and neutral. We compared two things in the experiment: using only the lexical of sentiment word; and using the score of the sentiment word.

TABLE 1. Experimental Result on Usage of Translated SentiWordNet

| Algorithm | Lexical value | Sentiment Score |
|---|---|---|
| Naïve bayes | 73.1% | 77.4% |
| Maximum Entropy | 73.2% | 78.4% |
| Support Vector Machine | 74.3% | 77.8% |

The experimental results on Table 1 strengthened other research results on the usage of SentiWordNet which showed that using sentiment score in the classification gave higher accuracy than only using the lexical words. The sentiment score can differentiate the word with low score of certain sentiment and the word with high score of sentiment. By using sentiment score, word with low sentiment score might have been ignored and give result of neutral or opposite class, while for the lexical value, word listed in the vocabulary can't be ignored and may give incorrect result.

For example, in the text "Berlibur di Hanoi Vietnam juga bisa menjadi liburan yang berbeda saat Tahun Baru Cina nanti Travelers"(Vacation at Hanoi, Vietnam can be a different vacation while Chinese New Year), the word "bisa" (can) has a sentiment score of 0.125 which shows a low possitiove sentiment. By using the sentiment score and term weight, the word "bisa" is ignored and it gives the neutral sentiment, which is a correct one. But, by using only the lexical value, the word "bisa" can't be ignored and it gives the possitive sentiment, which is an incorrect result.

*2) Experiments on classification method:*

The second experiment evaluated the direct and leveled method in sentimen classification. Both of the classifcation use the sentiment score feature as one of the base feature. The results in Table 2 showed that direct classification gave higher accuracy than leveled method. Both of the methods use the same feature. Here are the list of the feature that used by both method:

1. Unigram
2. Sentimen Score
3. Question words.

TABLE 2. Experimental Result on Direct and Leveled Classification

| Algorithm | Leveled | Direct |
|---|---|---|
| Naïve bayes | 76.5% | 77.4% |
| Maximum Entropy | 76.7% | 78.4% |
| Support Vector Machine | 77.3% | 77.8% |

able 2 shows that the leveled method give lower result in sentiment classification. If we break down the classfication, the positive and negative classification give more than 95% accuracy. The problem lies in the opinion and neutral classification that only give 78% accuracy. In order to improve the accuracy, more feature to detect neutral text is required.

*3) Experiments on sarcasm detection.*

The third experiment evaluated the additional features of negativity and interjection number in the sarcasm classification accuracy. The results in Table 3 showed that the additional features are effective in sarcasm detection.

TABLE 3. Experimental Result on Usage of Negativity and Interjection for Detecting Sarcasm

| Algorithm | Unigram | Unigram + Negativity + Interjection |
|---|---|---|
| Naïve bayes | 45.7% | 53.1% |
| Maximum Entropy | 47.1% | 53.8% |
| Support Vector Machine | 48.5% | 54.1% |

The additional features also showed that Indonesian people tend to write their critics using sarcasm. As for the low accuracy, we found that there are many sarcasm texts have no global topic. For example, in the text "Men, lu ganteng banget kalo pake dress. :p" (Man, you are so handsome using dress). Here, the text topic is not widely known, thus, the negativity feature is useless.

IV. CONCLUSIONS

We have shown that the additional features for detecting sarcasm are quite effective since it increased the accuracy of 6%. Based on our observation of the sarcasm data, we added the features of negativity and number of interjection words. The negativity feature tried to catch the global sentiment value, while the interjection feature represents the lexical phenomena of the text message. In our next research, we will evaluate the method to detect the sarcasm text without global topic and adding some new features to leveled classication to improve the result.